\documentclass{article}
\usepackage{spconf,amsmath,graphicx}

\title{SBP-Net: Learning Thin Structure Reconstruction with Sliding-Box Projections}
\name{Ofir Gilad, Andrei Sharf}
\address{Faculty of Computer and Information Science, Ben Gurion University of the Negev\\
Beer Sheva, Israel}

\usepackage{amssymb}
\usepackage{amsmath}

\usepackage{algorithm}
\usepackage{algorithmic}
\usepackage{url}

\usepackage{eso-pic}
\makeatletter
\def\copyrightnotice#1{%
  \gdef\@copyrightnotice{#1}%
  \AddToShipoutPictureBG*{%
    \AtPageLowerLeft{%
      \raisebox{1.8\baselineskip}{%
        \hspace{\dimexpr0.5\paperwidth-0.5\textwidth\relax}%
        \parbox{\textwidth}{\footnotesize\@copyrightnotice}%
      }%
    }%
  }%
}
\makeatother

\begin{document}
\maketitle

\copyrightnotice{\copyright\ 2026 IEEE. Personal use of this material is permitted. Permission from IEEE must be obtained for all other uses, in any current or future media, including reprinting/republishing this material for advertising or promotional purposes, creating new collective works, for resale or redistribution to servers or lists, or reuse of any copyrighted component of this work in other works.}

\begin{abstract}
Reconstructing thin 3D structures is challenging due to their sparsity, scale variation, and complex geometry. Such structures arise in a wide range of domains, including medical imaging of vascular systems and industrial pipe systems. While recent neural methods perform well on dense surfaces, they often fail to recover fine thin geometries. We propose a reconstruction approach based on local depth projections, which provide an efficient and informative 2D representation of thin structures. Specifically, we traverse the 3D model with a sliding box to generate local orthographic depth projections, which are processed by a neural network to reconstruct missing thin structures in 2D. The local reconstructions are subsequently fused back into the 3D model to produce a coherent and detailed shape. Experiments on pulmonary artery reconstruction from CT volumes and industrial pipeline recovery from synthetic and real scans demonstrate improved preservation of fine structural details over existing methods.
\end{abstract}

\section{Introduction}
\label{sec:intro}

\begin{figure}[t]
    \centering
    \includegraphics[width=0.76\linewidth]{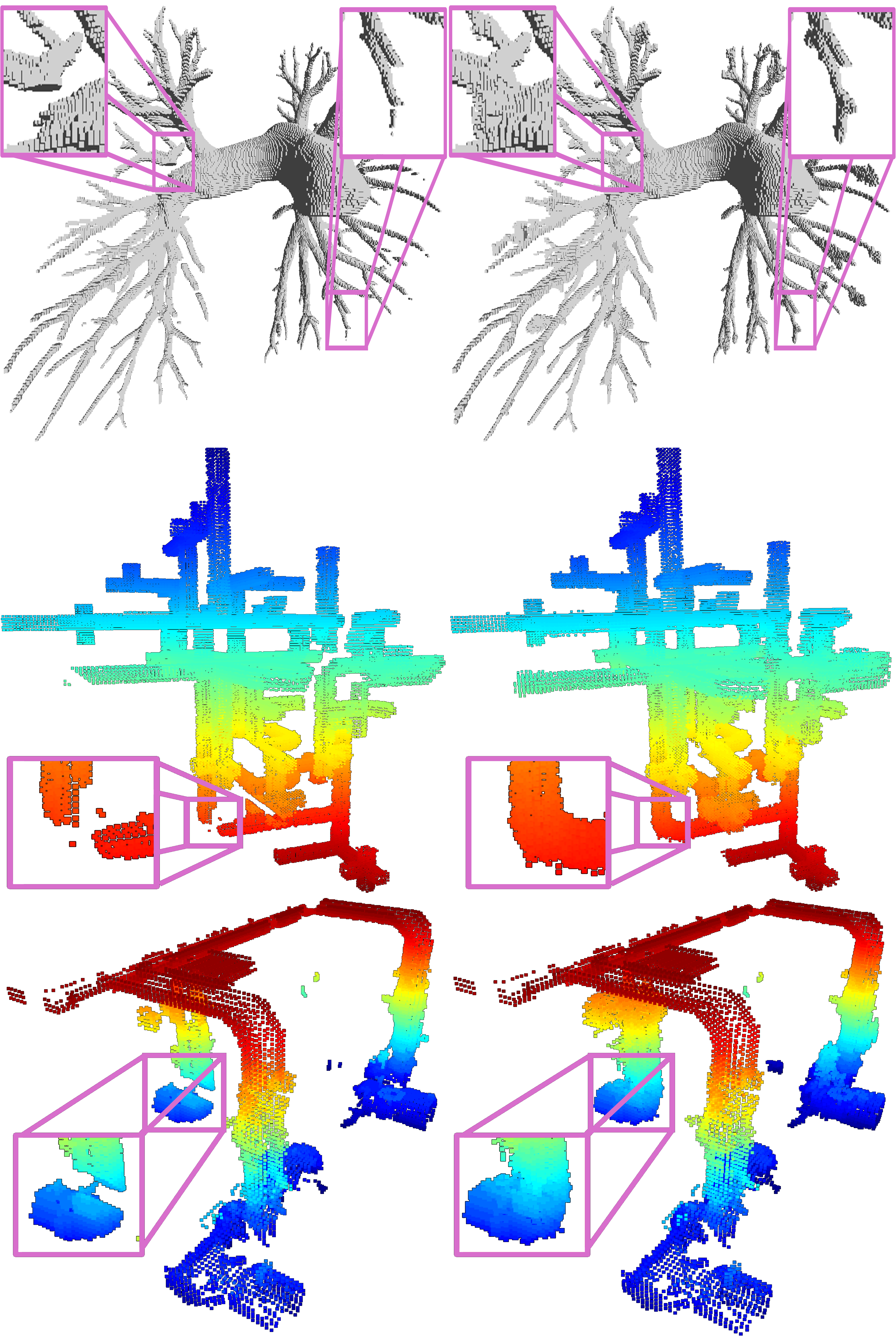}
    \caption{
    SBP-Net reconstruction of thin 3D structures on a pulmonary artery CT volume (top row), a synthetical power-plant scanned point cloud (mid row), and a real raw scan of pipes structure in a hospital (right row). 
    Left is input, and right is our result (zoom-in shows sliding-box sample).
    }
    \label{fig:Parse2022_3D}
\end{figure}

Reconstructing 3D objects with thin and intricate structures—such as vascular networks, neural tissues, or industrial pipelines—remains a fundamental challenge in computer vision. These objects are characterized by extreme sparsity, complex branching geometry, and large spatial extent, making them particularly sensitive to missing data, noise, and limited viewpoints. Accurately recovering both fine-scale geometry and global connectivity is critical in applications ranging from medical imaging to industrial inspection.

Classical 3D reconstruction methods often fail in this task: surface-based approaches require dense sampling, while volumetric methods demand prohibitively high resolution to preserve thin structures, frequently resulting in over-smoothing and topological errors.

Recent neural reconstruction techniques have significantly improved 3D modeling, yet limitations remain. Neural Radiance Fields (NeRF)~\cite{Mildenhall21} represent scenes as continuous volumetric fields but rely on density thresholding, which hinders accurate surface recovery for thin or sparse geometry. Implicit surface representations, including occupancy networks~\cite{Oechsle2021} and Signed Distance Functions (SDFs)~\cite{Park_2019}, improve surface quality, with volumetric re-parameterizations further enhancing rendering and geometry~\cite{Yariv2021VolumeRO,10.5555/3540261.3542342,Yariv2020MultiviewNS}. 

Despite these advances, such methods remain sensitive to sparse or low-quality input and often fail to preserve delicate branching structures and topological continuity.

Neural shape completion methods address missing data using learned priors over representations such as voxels~\cite{Wu20143DSA,Dai2016ShapeCU}, images~\cite{Jiajun18}, and point clouds~\cite{Stutz2018Learning3S}. While effective for compact objects, their reliance on global latent encodings limits scalability to large, thin, and highly connected structures. 

In medical imaging, thin anatomical structures are often observed through sparse cross-sectional scans. Early reconstruction methods primarily focused on parallel slice acquisition~\cite{barequet1994piecewise,bajaj1996arbitrary,boissonnat2007shape}, whereas more recent neural approaches address non-parallel acquisition settings~\cite{OReX2023,walker2024crosssdf}.

However, methods such as CrossSDF~\cite{walker2024crosssdf} and VesselSDF~\cite{EspSal_VesselSDF_MICCAI2025} typically assume watertight geometries and aim to reconstruct a global signed distance field, which limits their performance when applied to incomplete or imperfect structures.
In contrast, skeleton-based completion methods~\cite{Ren_Selfsupervised_MICCAI2024} focus on recovering topological centerlines, but do not reconstruct the full volumetric geometry.
A recent projection-based method~\cite{wang2025deepca} is related to our approach but does not explicitly address topological preservation in thin structures.

To address these challenges, we propose a projection-based framework for reconstructing imperfect thin 3D structures using local depth projections.
Our method employs a sliding-box depth projection operator (SBP) from six viewpoints, converting the global 3D reconstruction task into a set of structured 2D representations that are easier to complete and regularize. These projections are processed using an attention-enhanced encoder–decoder network that captures long-range dependencies to preserve connectivity across missing regions. A dedicated fusion module then integrates the completed projections into a coherent and spatially consistent 3D reconstruction.

We demonstrate the effectiveness of our approach on two real-world applications: reconstructing pulmonary arteries from CT volumes and modeling complex industrial pipeline networks from 3D scans. The results show that our method robustly preserves thin structures and topological continuity under sparse and imperfect supervision (see Figure~\ref{fig:Parse2022_3D}).

\section{Method}

\begin{figure}[t]
    \centering
    \includegraphics[width=1.0\linewidth]{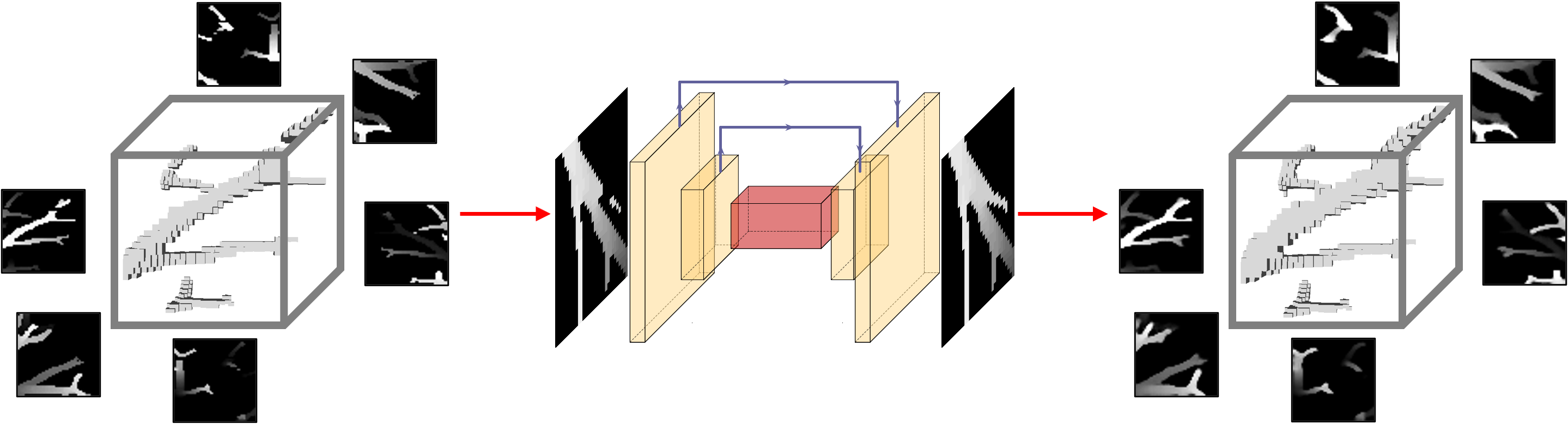}\\
    \caption{
    Pipeline illustration of SBP-Net pipeline. 
    Left-to-right, orthographic depth projections are fed to a 2D reconstruction U-Net. The reconstructed data is then reprojected and fused back in 3D.
    }  
    \label{fig:Flow}
\end{figure}

Our method reconstructs imperfect thin 3D structures by first transforming the problem from 3D to 2D using a sliding-box projection (SBP) operator. This approach simplifies the global 3D reconstruction into structured 2D representations, making them easier to complete and regularize. We then apply a custom attention U-Net model for the 2D reconstruction and reproject the solution back into 3D (see Figure~\ref{fig:Flow}). We discuss these details in the following.

\textbf{Sliding-box traversal and ortho-depth projection.}
Our method takes as input 3D objects represented by a volumetric grid.
For CT segmentation volumes, this grid is inherently defined by the voxel-based data representation, while for other data types, such as meshes or point clouds, we normalize and scale the input to fit onto canonical $1 \times 1 \times 1$ bounding box cells, and convert it into a voxel-based representation.
Thus, our voxel grid representation allows our model to efficiently project and learn local geometry completion using 2D convolutions as follows. 

Next, a 3D sliding-box operator traverses the grid to extract local sub-volume segments. Using a fixed-size sliding-box with a step size equal to half its edge length, we ensure that neighboring boxes overlap, thus promoting continuity during subsequent reconstruction and fusion stages.

For each segment, we compute orthographic 2D depth projections from six canonical viewpoints: \textit{top}, \textit{bottom}, \textit{left}, \textit{right}, \textit{front}, and \textit{back}. 
Each projection acts as a Z-buffer, recording the minimum per-pixel depth value $d$ along the corresponding viewing direction. Then, we normalize these values to a $0-255$ intensity range: $\left\lfloor \left(1 - \frac{d}{N - 1}\right) \cdot 255 \right\rceil$, where $N$ is the box size. This yields 6 independent orthogonal 1-channel grayscale images, which are fed arbitrarily to the network during training.

\textbf{2D reconstruction network.}
The orthographic depth projections serve as input to a lightweight neural network designed to reconstruct complete 2D structures by filling in missing regions. Our model is a custom variant of the Attention U-Net architecture \cite{oktay2018attentionunetlearninglook}, comprising four convolutional encoder blocks connected via skip connections to four transposed convolutional decoder blocks, with a single self-attention module incorporated at the bottleneck. ReLU activations are applied throughout the network, and a final Sigmoid activation maps the output to $[0, 1]$ grayscale range.

Skip connections preserve the original input and enhance feature coherence across the network, while the self-attention module enables consistent and accurate hole-filling by capturing global contextual information. Together, these components allow the model to effectively reconstruct missing regions in the 2D projections, providing reliable inputs for the subsequent 3D reconstruction stage.

\textbf{Re-projection and data fusion.}
The refined 2D projections are re-projected and integrated back into 3D. In this step, each depth image is mapped back into its corresponding 3D sub-volume, effectively reversing the 2D depth projection. Once all sub-volumes are reconstructed, they are merged and fused with the original 3D input using a logical \texttt{OR} operation, ensuring that only newly reconstructed content is added. 

Because the localized sliding-boxes decompose complex global structures into simple geometric sub-shapes, this straightforward operator safely restricts voxel addition to previously empty regions, preventing over-reconstruction and preserving the original structural integrity.

Finally, all completed segment reconstructions are reintegrated, producing a coherent and structurally complete 3D model. This process ensures continuity across segments while maintaining the fidelity of the original data.

\section{Optimization}

\textbf{Training data.}
Given the 3D ground truth $X$ and an incomplete input segment $X'$, we compute a binary difference map $D = \text{bin}(|X - X'| > 0)$ where each nonzero region in (D) is treated as a potential hole and represented as an isolated connected component $d_i \in D$.

To distinguish true holes from naturally occurring narrow gaps, we assess whether completing a potential hole would alter the assumed tree-like topology of the thin structures (i.e., create a loop). Specifically, we evaluate whether filling a hole in $X'$ would change the topological connectivity of $X$. This ensures that only  holes are completed, avoiding the merging of distinct connected components in the original structure.

For each candidate component $d_i$, we define a tight local neighborhood using a binary mask $M$, created by dilating the smallest bounding box that encloses $d_i$. Within this neighborhood, we count the number of connected components in the input before filling $(n_{\text{base}})$ and after filling the gap $(n_{\text{new}})$. If $n_{\text{new}} = n_{\text{base}}$, meaning no new structural connections are introduced, the gap is considered insignificant and is filled. Conversely, if $n_{\text{new}} < n_{\text{base}}$, the gap is preserved as a meaningful structural hole.

\textbf{Training loss.}
The 2D regression model is trained on the refined data using L1 loss with a weighted mask and Adam optimizer. This encourages the network to fill meaningful holes while preserving existing structures and background.

Let \(P_v\) denote the orthographic projection from view $v$.
The loss is defined based on the input projection \(P_v'\) and the ground truth projection \(P_v\):

\[
M_{\text{fill}} = \text{bin}((P_v - P_v') > 0) \quad \text{(hole filling)}
\]
\[
M_{\text{bg}} = \text{bin}(P_v = 0) \quad \text{(preserve background)}
\]
\[
M_{\text{fg}} = 1 - (M_{\text{fill}} + M_{\text{bg}}) \quad \text{(preserve foreground)}
\]
\[
M_{\text{w}} = w_{\text{fill}} \cdot M_{\text{fill}} + w_{\text{fg}} \cdot M_{\text{fg}} + w_{\text{bg}} \cdot M_{\text{bg}}
\]
where $w_{\text{fill}} = 0.85, w_{\text{fg}} = 0.10, w_{\text{bg}} = 0.05$.

The final loss is then defined as $
\mathcal{L}_{\text{mask}} = \lambda_{\text{mask}} \cdot M_{\text{w}} \odot |P_v' - P_v|
$ where $\lambda_{\text{mask}} = 100$ is a scaling factor to amplify the loss signal.

This mask balances the model’s focus between hole filling and preserving structural integrity, enabling rapid convergence and high-quality results with limited training data \cite{li2020recurrentfeaturereasoningimage}.

\section{Experiments and Results}

To evaluate our method, we conducted experiments on three datasets focused on thin-structures: medical CT scans of pulmonary arteries (\textit{PARSE 2022}), synthetic 3D models of industrial pipes (\textit{PipeForge3D}), and real-world 3D point clouds (\textit{Hospital Central Utility Plant}). In Figure~\ref{fig:Parse2022_3D}, rows show results for each dataset respectively. Code is available at \url{https://github.com/OfirGiladBGU/SBP_Net}.

\textbf{Pulmonary Artery CT Dataset (PARSE).}
We evaluate our method on the \textit{PARSE} benchmark from the Pulmonary Artery Segmentation Challenge~\cite{luo2023efficient}, which contains 200 CT volumes acquired with a dual-source 64-slice scanner and voxel-level pulmonary artery annotations provided by ten clinicians. Volumes have resolutions ranging from $512 \times 512 \times 228$ to $512 \times 512 \times 376$, with in-plane spacing of 0.50–0.95\,mm and a slice thickness of 1\,mm. Voxel labels are binary, with 0 for background and 1 for arteries, making the dataset suited for thin vascular reconstruction evaluation.

We train our model on 50 annotated CT scans. To focus on fine peripheral branches, we first apply a state-of-the-art segmentation method~\cite{carmo2024medpseghierarchicalpolymorphicmultitask} to extract large vessels. Training regions are then identified by comparing these results with the ground truth and selecting regions of discrepancy corresponding to thin or missing arterial branches. A sliding-box traversal is applied to these regions in both input and ground truth volumes to generate training samples.

\textbf{Synthetic Pipes Dataset (PipeForge3D).}
To evaluate generalization across domains, we construct a synthetic pipe dataset using \emph{PipeForge3D}~\cite{pipeforge3d}, a custom automatic 3D pipe generator that produces models in both mesh and point cloud formats by assembling predefined pipe components. Each model is generated by recursively constructing a random 3D graph \( G = (V, E) \), where nodes select random right-angle connectors defining edges along the X, Y, and Z axes. Mesh connectors are placed at node locations, edges are replaced with tubular structures, and the resulting triangular mesh is sampled with a scanning simulator to obtain point clouds. Using this process, we generate 50 industrial pipe models, each containing approximately 100 pipe components.

To simulate missing thin structures, we duplicate the ground-truth data and randomly remove regions, creating on average ten holes per sample. Each hole is formed by selecting a random plane and removing voxels within a thin aligned box. For point cloud data, we assume sufficiently dense sampling to distinguish true structural holes from sparsely sampled regions after 2D projection.

\textbf{Power Plant Real Scans Dataset.}
We also evaluate our method on real-world industrial power plant scans captured with ClearEdge3D’s high-precision laser scanning technology~\cite{clearedge3d}, represented as point clouds. We crop and voxelize regions containing thin pipe structures for evaluation. As ground truth is unavailable, the model is pre-trained on synthetic data and evaluated by injecting synthetic holes into the real scans and measuring reconstruction accuracy.

Note that even for raw real-world scans of industrial pipes, our model effectively detects and completes missing or damaged regions despite noise and structural complexity typical of such data (see Figure~\ref{fig:Parse2022_3D} (bottom)).

\textbf{Metrics.}
We evaluate hole completion using metrics capturing both local accuracy and global fidelity. MAE$^{mask}$ and RMSE$^{mask}$ measure average and large depth deviations in masked regions, respectively. SSIM assesses geometric similarity between predicted and ground-truth depths, $\delta_1^{mask}$ quantifies relative accuracy within 1.25× of ground truth~\cite{6909413}, and Dice measures volumetric overlap, reflecting shape consistency.
For thin structure reconstruction, we compare against baselines using Chamfer Distance (CD) for overall geometric accuracy, Hausdorff Distance (HD) for worst-case deviations, and Connected Components (CC) count to assess topological coherence, where a single component (26-connectivity) indicates structural correctness.

\begin{figure}[t]
    \centering
    \includegraphics[width=1.0\linewidth]{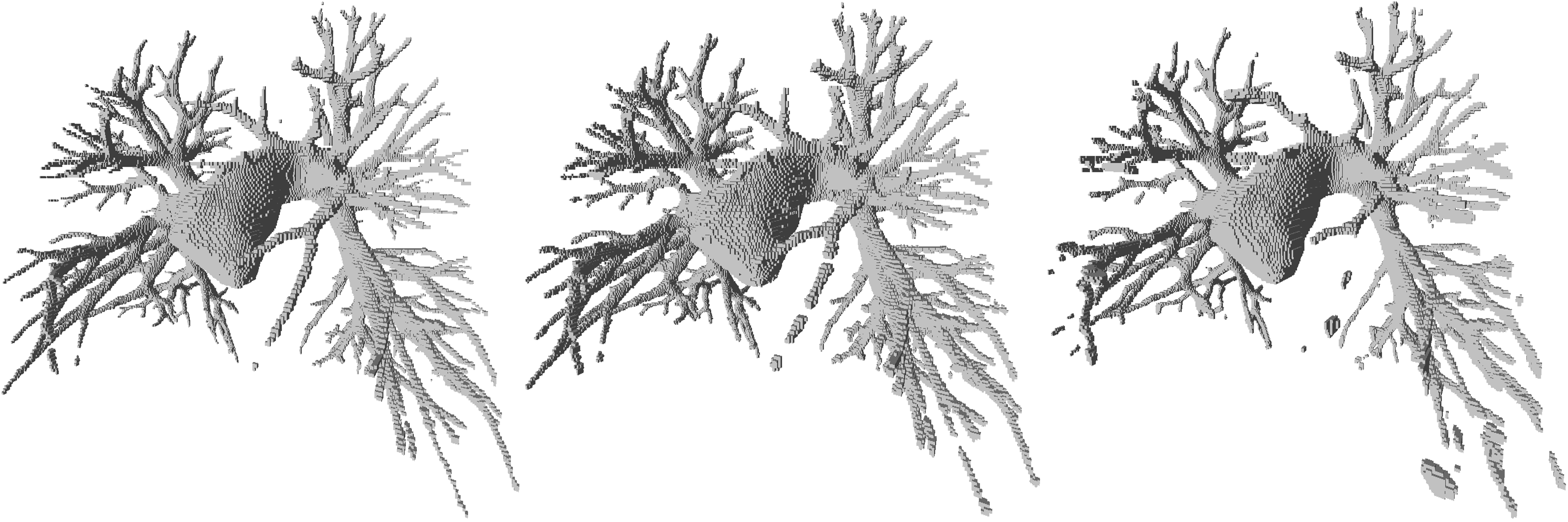}\\
    \centerline{(a)\hspace{0.29\linewidth}(b)\hspace{0.29\linewidth}(c)}
    \centering
    \includegraphics[width=1.0\linewidth]{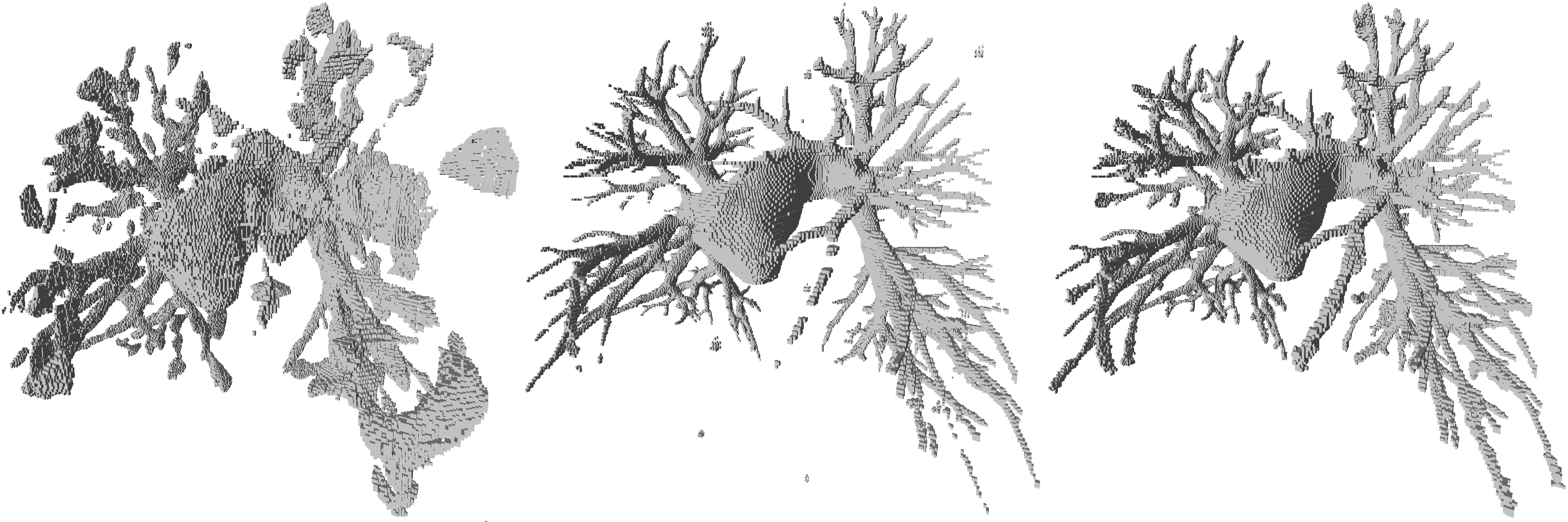}\\
    \centerline{(d)\hspace{0.29\linewidth}(e)\hspace{0.29\linewidth}(f)}
    \caption{
     Comparisons on reconstruction results on PARSE thin structures. (a) UNet3D , (b) 3D-RecGAN, (c) Conv ONet, (d) OReX, (e) DeepCA, and (f) SBP-Net.
    }
    \label{fig:SOTA_Compare}
\end{figure}

\textbf{Evaluation and Comparisons.}
Figure~\ref{fig:SOTA_Compare} presents qualitative comparisons on  \textit{PARSE} thin Pulmonary Arteries.
Our method achieves superior reconstruction of thin structures and a more complete recovery of missing regions than baselines.

\begin{figure}[t]
    \centering
    \includegraphics[width=0.67\linewidth]{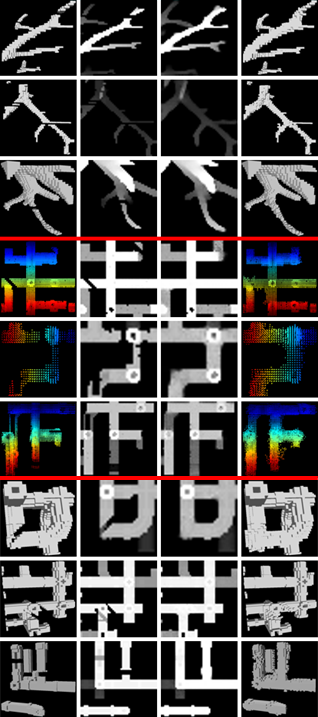}
    \centerline{(a)\hspace{0.12\linewidth}(b)\hspace{0.12\linewidth}(c)\hspace{0.12\linewidth}(d)}
    \caption{
    Reconstruction results:  arteries CT volumes (rows 1-3), 3D pipes point clouds (rows 4-6), and 3D pipes voxel grid meshes (rows 7-9). 
    Columns are sliding-box inputs with holes (a), 2D depth projection (b), our 2D reconstruction (c), and our reconstructed 3D volume (d).
    }
    \label{fig:Combined}
\end{figure}

Figure~\ref{fig:Combined} presents additional qualitative results on selected sliding-box samples from the \textit{PARSE} and \textit{PipeForge3D} datasets. The \textit{PARSE} dataset is given as CT volume grid while \textit{PipeForge3D} is given both as discretely sampled point clouds as well as a continuous mesh that is voxelized. For each sliding-box, we display the corresponding 3D sub-volume segments together with their 2D projections from selected viewpoints and the reconstructed results.

Our model simultaneously detects and fills multiple holes while preserving the topology of peripheral branches in the pulmonary dataset (rows 1–3).
It also performs well on discrete point cloud representations (rows 4–6), effectively distinguishing between true missing regions and sparse sampling.
On voxelized synthetic pipe meshes (rows 7–9), the model accurately identifies and fills holes even in complex, self-occluded configurations.

Table~\ref{table:1} summarizes our model’s accuracy across multiple error metrics on four datasets: \textit{PARSE}, \textit{PipeForge3D} voxelized mesh, \textit{PipeForge3D} scanned point cloud, and the \textit{Hospital} Power Plant real scans.
The CT \textit{PARSE} data and the synthetic pipe datasets—both continuous mesh and point cloud representations—achieve similarly high accuracy, while performance on real raw scans is lower due to substantial noise and extensive missing regions.

\begin{table}[t]
\centering
\begin{tabular}
 { |c||c|c|c|c| }
 \hline
  Dataset & PARSE    & Pipe  & Pipe  & Hospital \\
          & 2022     & Forge & Forge & CUP \\
          & (CT seg) & (vol) & (pts) & (scans)  \\
 \hline
 MAE$^{mask}$       & 0.072 & 0.106 & 0.092 & 0.233 \\
 RMSE$^{mask}$      & 0.121 & 0.130 & 0.117 & 0.284 \\
 SSIM               & 0.811 & 0.989 & 0.984 & 0.854 \\
 $\delta_1^{mask}$  & 0.812 & 0.736 & 0.731 & 0.425 \\
 Dice               & 0.930 & 0.983 & 0.963 & 0.840 \\
 \hline
\end{tabular}
\caption{
Our method accuracy evaluation  on four datasets: PARSE  CT, PipeForge as voxelized mesh, PipeForge as point clouds, and real scans of a Hospital power plant.
Evaluation metrics are: MAE, RMSE, $\delta_1$ in the missing region, SSIM and Dice for all data.
}
\label{table:1}
\end{table}

\begin{table}[t]
\centering
\begin{tabular}
 { |c||c|c|c|c| }
 \hline
 Metrics       & CD     & HD     & CC  \\
 \hline
 SBP-Net       & 0.446 & \textbf{17.899} & \textbf{2} \\
 DeepCA        & 3.150  & 113.871 & 212 \\
 OReX          & 21.323 & 55.246  & 298 \\
 Conv ONet     &  2.687 & 22.857  & 203 \\
 3D-RecGAN     &  0.351 & 18.330  & 43  \\
 UNet3D        & \textbf{0.296} & 18.951  & 26  \\
 \hline
\end{tabular}
\caption{
Comparisons of our SBP-Net method with DeepCA, OReX, Conv ONet, 3D-RecGAN, and UNet3D.
Reconstructed outputs are evaluated using Chamfer Distance (CD), Hausdorff Distance (HD), and Connected Components (CC) relative to the ground truth.
}
\label{table:2}
\end{table}

\begin{table}[t]
\centering
\begin{tabular}
 { |c||c|c|c| }
 \hline
 Box Size           & $32^3$ & $48^3$ & $64^3$ \\
 \hline
 MAE$^{mask}$       & 0.072 & 0.075 & 0.085 \\
 RMSE$^{mask}$      & 0.121 & 0.136 & 0.154 \\
 SSIM               & 0.811 & 0.803 & 0.832 \\
 $\delta_1^{mask}$  & 0.812 & 0.798 & 0.766 \\
 Dice               & 0.930 & 0.940 & 0.940 \\
 \hline
\end{tabular}
\caption{
Evaluation of local box size impact on reconstruction accuracy.
Columns report our method’s performance across four error metrics for different sliding-box sizes.
}
\label{table:3}
\end{table}

\begin{table}[t]
\centering
\begin{tabular}
 { |c||c|c|c|c| }
 \hline
 Avg. Hole Size  & $125$ & $250$ & $450$ & $750$ \\
 \hline
 MAE$^{mask}$       & 0.095 & 0.102 & 0.115 & 0.125 \\
 RMSE$^{mask}$      & 0.122 & 0.132 & 0.141 & 0.158 \\
 SSIM               & 0.989 & 0.978 & 0.950 & 0.851 \\
 $\delta_1^{mask}$  & 0.751 & 0.748 & 0.695 & 0.687 \\
 Dice               & 0.980 & 0.973 & 0.957 & 0.886 \\
 \hline
\end{tabular}
\caption{
Evaluation of our method’s performance with respect to missing region size.
Columns show results across four error metrics for different missing region sizes.
}
\label{table:4}
\end{table}

We compare our method, on \textit{PARSE} dataset, with related SOTA 3D reconstruction methods: DeepCA~\cite{wang2025deepca}, OReX~\cite{OReX2023} (with 100 slices), Convolutional Occupancy Networks~\cite{Peng2020ECCV}, 3D-RecGAN~\cite{Yang17} and UNet3D~\cite{cicek20163dunetlearningdense} 
(See Figures~\ref{fig:SOTA_Compare}).

Note that our method demonstrates higher-quality reconstruction, accurately filling all missing regions while preserving the original structure and avoiding false-positive completions that introduce noise.

Table~\ref{table:2} summarizes this comparison results using Dice Score, Chamfer Distance (CD), Hausdorff Distance (HD), and Connected Components (CC), all computed relative to the ground truth.

Thanks to its strong topological component in completing holes, our method significantly outperforms others in CC metric as well as in HD. However, in Chamfer Distance, our method ranks third behind UNet3D and 3D-RecGAN, as focusing on accurately filling missing regions can introduce minor noise in the overall shape.

\textbf{Ablation Study}
Table~\ref{table:3} illustrates the impact of sliding-box size on reconstruction accuracy.
We evaluated our model on the \textit{PARSE} dataset using varying box sizes. As box size increases, errors grow across all metrics because larger boxes produce projections with more occlusions, reducing their informativeness. Nonetheless, the moderate increase in error demonstrates the robustness of our method.

Table~\ref{table:4} shows the effect of hole size on reconstruction accuracy.
Using synthetic pipes from \textit{PipeForge3D}, we introduced random holes of varying size, defined by the average number of missing voxels per hole. Holes were generated as described, by intersecting plane bands with increasing width. As expected, reconstruction error grows with hole size moderately, further demonstrating the robustness of our method.

\textbf{Limitations.}
This work tackles the challenge of detecting and reconstructing incomplete thin 3D structures using depth projections as a compact, informative 2D representation of local geometry. The approach achieves strong performance across both medical and industrial applications.
Nevertheless, several limitations remain.
Although rare due to the locality of our sliding-box operator, self-occluded holes hidden behind foreground geometry in certain orthographic views are particularly challenging to detect and reconstruct.
Relying on 2D grayscale projections introduces a strong inductive bias, which can limit generalization to novel structures or unseen domains.
Future work could explore more robust 3D-aware representations, multiview consistency, and explicit occlusion reasoning to overcome these challenges.

\bibliographystyle{IEEEbib}
\bibliography{strings,research_fixed}

\end{document}